\documentclass{article}

\PassOptionsToPackage{numbers, compress}{natbib}

\usepackage[preprint]{neurips_2020}

\usepackage[utf8]{inputenc} 
\usepackage[T1]{fontenc}    
\usepackage{hyperref}       
\usepackage{url}            
\usepackage{booktabs}       
\usepackage{amsfonts}       
\usepackage{nicefrac}       
\usepackage{microtype}      
\usepackage{amsmath}
\usepackage{graphicx}
\usepackage[linesnumbered,ruled]{algorithm2e}
\usepackage{amsthm}

\title{Safety-guaranteed Reinforcement Learning based on Multi-class Support Vector Machine}

\author{
  Kwangyeon Kim \\
  Purdue University\\
  West Lafayette, IN \\
  \texttt{kim1601@purdue.edu} \\
  \And
  Akshita Gupta \\
  Purdue University\\
  West Lafayette, IN \\
  \texttt{gupta417@purdue.edu} \\
  \AND
  Hong-Cheol Choi \\
  Purdue University\\
  West Lafayette, IN \\
  \texttt{choi642@purdue.edu} \\
  \And
  Inseok Hwang \\
  Purdue University\\
  West Lafayette, IN \\
  \texttt{ihwang@purdue.edu} \\
}

\begin{document}

\maketitle

\begin{abstract}
  Several works have addressed the problem of incorporating constraints in the reinforcement learning (RL) framework, however majority of them can only guarantee the satisfaction of soft constraints. In this work, we address the problem of satisfying hard state constraints in a model-free RL setting with the deterministic system dynamics. The proposed algorithm is developed for the discrete state and action space and utilizes a multi-class support vector machine (SVM) to represent the policy. The state constraints are incorporated in the SVM optimization framework to derive an analytical solution for determining the policy parameters. This final policy converges to a solution which is guaranteed to satisfy the constraints. Additionally, the proposed formulation adheres to the Q-learning framework and thus, also guarantees convergence to the optimal solution. The algorithm is demonstrated with multiple example problems.
\end{abstract}


\section{Introduction}
The growing success of reinforcement learning (RL) algorithms in utilizing data to derive optimal controllers for complex gaming environments \cite{ mnih2013playing, schraudolph1994temporal, silver2016mastering, tesauro1994td} has prompted researches to test its application to real world problems as well.
In particular, physical systems such as robotic systems pose a challenging environment because the RL agent needs to accurately learn the optimal behavior, as well as it also has to satisfy safety constraints on the system to avoid any damage to hardware \cite{kormushev2013reinforcement, polydoros2017survey}.
Despite these difficulties, RL shows potential in tackling problems such as automotive driving \cite{sallab2017deep, chen2019model}, controlling humanoid robots \cite{peters2003reinforcement, ilg1995learning, stulp2010reinforcement} and learning motor skills for industrial robots \cite{kormushev2010robot, gullapalli1994acquiring, gu2017deep}.
However, incorporating constraints in the RL framework is still a challenge.
In this paper, we propose an algorithm that can incorporate the safety constraints in the RL framework and ensures the safety during learning and execution.

Moldovan and Abbeel \cite{moldovan2012safe} highlighted the importance of constrained RL in their work by stating that most physical systems do not satisfy the widely used assumption of ergodicity.
That is, in most physical systems, all states cannot be reached from every possible state.
This implies that if a system enters an unsafe state, there may not exist a policy which brings it back to a safe state.
With the increasing need of deployment of RL algorithms on real systems, the above statement directly links the unsafe exploration phase of RL algorithms to compromising the hardware of real systems.
Thus, in the last few decades, researches have gravitated towards the problem of incorporating constraints in the RL framework. 

Garcia and Fernandez \cite{garcia2015comprehensive} classified constrained RL algorithms based on whether they were modifying the optimization criteria of the agent \cite{moldovan2012safe, kadota2006discounted, ng1999policy} or they were modifying the exploration process by incorporating external guidance \cite{quintia2013learning, torrey2012help, driessens2004integrating}.
However, the former methods can guarantee the satisfaction of soft constraints only.
To ensure that the RL agent satisfies hard constraints imposed on the state space, additional information is required to prevent the agent from entering a constrained, unsafe region.
For this purpose, some works proposed switching between base-level controllers \cite{perkins2002lyapunov, fisac2018general, chow2018lyapunov} which are guaranteed to be safe while offering satisfactory behavior in terms of optimizing the reward function.
However, some of these works made rather strong assumptions of starting with known base-level safe policies \cite{perkins2002lyapunov, chow2018lyapunov} or knowing the Lyapunov function of the system \cite{berkenkamp2017safe}.
Other approaches formulated the problem as a constrained Markov Decision Process (CMDP) \cite{chow2017risk, achiam2017constrained, junges2016safety}.
However, due to the nonlinear, non-convex nature of the policy function approximator, the heuristic-based gradient descent methods used to solve the optimization problem could not give any convergence guarantees on the solution in tractable time.

To overcome such challenges, the underlying idea of the proposed algorithm is to modify the policy update rule so that it converges to an optimal solution, while guaranteeing the safety during learning. To this end, we propose to use the Support Vector Machine (SVM) as a policy function in order to utilize the decent theoretical nature of it, since it is formulated as quadratic programming.

The rest of this paper is organized as follows: Section \ref{sec:2} formulates a safety-constrained RL problem. We then propose a safety-guaranteed algorithm using the SVM in Section \ref{sec:3}. The proposed algorithm is demonstrated with examples in Section \ref{sec:4}, and the conclusions and future works are presented in Section \ref{sec:5}.

\section{Problem Formulation} \label{sec:2}
Given the discrete state space $\mathcal{X} = \left\{ x^{(i)} \right\}_{i=1}^{N}$ and the discrete action space $\mathcal{U} = \left\{ u_{(k)} \right\}_{k=1}^{M}$, suppose an agent observes a state $x_t\in \mathcal{X}$ and take an action $u_t \in \mathcal{U}$, at discrete time step $t$. The environment then moves to the next state $x_{t+1}$ and returns the reward $r_{t+1}$, which are determined by the system dynamics:
\begin{align}
\begin{split}
    x_{t+1} &= f(x_t, u_t) \\
    r_{t+1} &= g(x_t, u_t)
\end{split}
\label{eqn:disc_dyn}
\end{align}
where $f:\mathcal{X} \times \mathcal{U} \rightarrow \mathcal{X}$ is the \textit{state transition function} (or system dynamics) and $g:\mathcal{X} \times \mathcal{U} \rightarrow \mathbb{R}$ is the \textit{reward function}, both of which are assumed to be deterministic. The agent is initialized at $x_0 \in \mathcal{X}_0 \subset \mathcal{X}$ at $t=0$ and terminates at $x_f \in \mathcal{X}_f \subset \mathcal{X}$ at $t=T$. The objective of the agent is to learn a \textit{policy function} $\pi:\mathcal{X} \rightarrow \mathcal{U}$ parameterized by $\theta$, denoted as $\pi(x; \theta)$, that (i) maximizes the return $R$ (the cumulative discounted reward) and (ii) satisfies the state constraint vector $\mathcal{C} (x_t) <0$ for all $t$ where $\mathcal{C}: \mathcal{X}\rightarrow \mathbb{R}^L$ with $L$ being the total number of constraints, which can be formally represented as: 
\begin{align}
\begin{split}
    \max R &= \sum_{t=0}^{T} \gamma^t r_{t+1}\\
    \text{s.t.\quad}\mathcal{C} (x_t) & < 0 \text{\quad for\quad} t=0,\cdots,T
\end{split}
\end{align}
where $\gamma \in (0,1]$ is a discount factor.

\section{Solution Approach} \label{sec:3}

To guarantee the safety during learning and execution, we assume that the environment dynamics is known a priori, that is, we consider a \textit{model-based} method, or \textit{planning} algorithm. For each state $x\in\mathcal{X}$, we define a \textit{set of safe actions}, $\mathcal{U}_{\text{safe}} (x)$:

\textbf{Definition 1} For a state $x\in\mathcal{X}$, assume that it is safe, i.e., $\mathcal{C}(x)<0$. A \textit{set of safe actions} for $x$ is defined as a set of actions $u\in\mathcal{U}$ such that $\mathcal{C} (f(x,u)) <0$, that is,
\begin{equation}
    \mathcal{U}_\text{safe} (x)=\left\{ u \in \mathcal{U} \ | \ \mathcal{C}(f(x,u)) <0\right\}
\end{equation}

\begin{figure}[h]
    \centering
    \graphicspath{./images/}
    \includegraphics[scale=0.4]{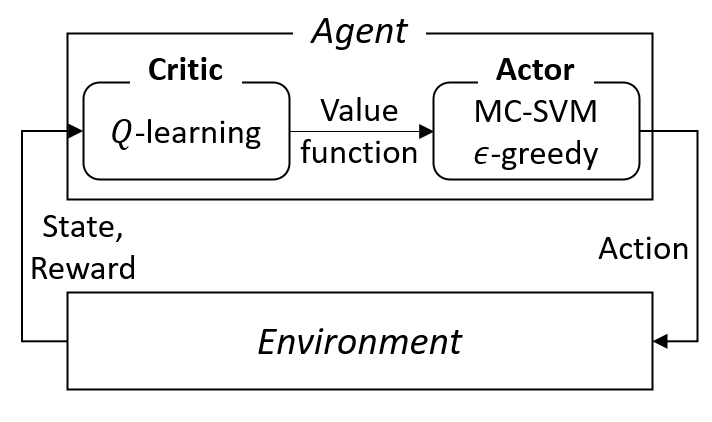}
    \caption{Proposed Actor-Critic framework using MC-SVM}
    \label{fig:AC_framework}
\end{figure}

We propose a safety-guaranteed reinforcement learning (RL) algorithm based on the Actor-Critic method \cite{konda2000actor}, in which the Critic evaluates the Actor's policy using a value function to guide the Actor to improve its policy, as shown in Fig.~\ref{fig:AC_framework}. 

At each time step $t$, the agent receives an observation of the environment, $x_t$, and takes an action $u_t$. The environment then returns the next state, $x_{t+1}$ and the reward, $r_{t+1}$. By taking $(x_t, u_t, x_{t+1}, r_{t+1})$, the Critic evaluates the current policy $\pi (x; \theta)$ to update its value function, by using $Q$-learning \cite{Watkins1992}: 
\begin{equation}
    Q(x_t, u_t) \leftarrow (1 - \beta) Q(x_t, u_t) + \beta \left( r_{t+1} + \gamma \max_{u \in \mathcal{U}_{\text{safe}}(x_{t+1})} Q(x_{t+1}, u)\right)
\label{eqn:q_update_svm}
\end{equation}
where $Q(x,u)$ is the state-action value function and $\beta \in [0,1]$ is the learning rate. Using the updated $Q$-table, i.e., $Q(x,u)$ for all $x\in\mathcal{X}$ and $u\in\mathcal{U}$, the Actor constructs a policy table in the form of a labeled dataset:
\begin{equation}
    \left\{ x^{(i)}, y^{(i)} =\arg \max_{u \in \mathcal{U}_{\text{safe}}(x^{(i)})} {Q\left( x^{(i)}, u \right)} \right\}_{i=1}^N
    \label{eqn:labeledData}
\end{equation}
The policy table means the current best guess of the optimal action for each state, among the safe actions. The Actor then updates its policy parameter $\theta$ so that the policy function $\pi (x;\theta)$ can represent the dataset as much as possible, which can be achieved by using a multi-class classification technique. To guarantee the safety as well as to obtain better provable properties, such as convergence and optimality, we use the Multi-Class Support Vector Machine (MC-SVM) which relies on solving a quadratic programming \cite{weston1998multi}. Given the labeled dataset $\left\{ x^{(i)}, y^{(i)} \right\}_{i=1}^N$, the MC-SVM is performed as follows:

\begin{itemize}
    \item For each $k\in\{1,\cdots,M\}$, a binary dataset $\Tilde{y}_{(k)}= \left\{\Tilde{y}_{(k)}^{(i)}\right\}_{i=1}^N$ is constructed for each $i=1,\cdots,N$ as:
    \begin{equation}
    \Tilde{y}_{(k)}^{(i)} = \begin{cases} +1 &\mbox{if } y^{(i)}=k \\
-1 & \mbox{if } y^{(i)}\neq k  \end{cases}
    \label{eqn:multilabels}
    \end{equation}
    \item We then solve a binary SVM optimization:
\begin{equation}
\begin{split}
    \min_{\alpha_{(k)}} W(\alpha_{(k)}) \; &= \; \frac{1}{2} \sum_{i=1}^{N} \sum_{j=1}^{N} \alpha_{(k)}^{(i)} \alpha_{(k)}^{(j)} \Tilde{y}_{(k)}^{(i)} \Tilde{y}_{(k)}^{(j)} K\left(x^{(i)}, x^{(j)}\right) - \sum_{i=1}^{N} \alpha_{(k)}^{(i)}  \\
    \text{subject to}& \sum_{i=1}^{N} \alpha_{(k)}^{(i)} \Tilde{y}_{(k)}^{(i)} = 0 \\
    & \alpha_{(k)}^{(i)} \in [0,U_b], \; \; \; \forall i = 1,2,\cdots ,N
\end{split}
\label{eqn:P2}
\end{equation}
where $\alpha_{(k)}= \left\{\alpha_{(k)}^{(i)}\right\}_{i=1}^N$ is the set of Lagrangian multipliers, $U_b$ is a regularization parameter for the trade-off between low classification error and high generalization performance, and $K\left(x^{(i)}, x^{(j)}\right)$ is a similarity measure between two states $x^{(i)}$ and $x^{(j)}$. We use the Gaussian kernel, which is widely used when there is no prior knowledge about the data: 
    \begin{equation}
        K\left(x^{(i)}, x^{(j)}\right) = \exp{(-\eta ||x^{(i)} - x^{(j)} ||^2)}
        \label{eqn"kernel_gauss}
    \end{equation}
    where $\eta$ is a design parameter.
    \item Using the computed Lagarangian multipliers, a binary decision-value function $h_{(k)} (x)$ is obtained as:
    \begin{equation}
    h_{(k)}(x) = \sum_{i=1}^N {\alpha_{(k)}^{(i)}}\Tilde{y}_{(k)}^{(i)} K\left(x, x^{(i)}\right) +b_{(k)}
    \label{eqn:binary_dec}
    \end{equation}
where $b_{(k)} \in \mathbb{R}$ is an offset term, computed by $b_{(k)} = \tilde{y}_{(k)}^{(j)} - \sum_{i=1}^N \alpha_{(k)}^{(i)} \tilde{y}_{(k)}^{(i)} K(x^{(i)}, x^{(j)})$ for any $x^{(j)}$ such that $\alpha^{(j)} \neq 0$.
    \item The policy function is then updated using the $M$ binary decision-value functions:
    \begin{equation}
        \pi \left(x; \theta \right) = \arg \max _{k\in\{1,\cdots,M\}} h_{(k)} \left(x\right)
        \label{eqn:policy_function}
    \end{equation}
\end{itemize}

We derive an analytical solution to determine the Lagrangian multipliers $\alpha_{(k)}$ for a binary SVM optimization. By developing a closed-form solution, we can bypass solving the optimization problem numerically, thereby reducing the computational time required to determine the binary SVM optimization. We define the sets $I_{(k)}^+ := \left\{ i \ | \ \tilde{y}_{(k)}^{(i)} = +1\right\}$ and $I_{(k)}^- := \left\{ i \ | \ \tilde{y}_{(k)}^{(i)} = -1\right\}$ to denote the sets of state indices corresponding to the positive and negative labels, respectively.

\textbf{Theorem 1} Let the Lagrangian multipliers $\alpha_{(k)}^{(i)}$ corresponding to $I_{(k)}^+$ and $I_{(k)}^-$ be $\alpha_{(k)}^+$ and $\alpha_{(k)}^-$, respectively. For $\eta\rightarrow\infty$, the analytic solution
\begin{equation}
\alpha_{(k)}^+ = \frac{2N_{(k)}^-}{N} \text{ and } \alpha_{(k)}^- = \frac{2N_{(k)}^+}{N}.
\label{eqn:analytic} 
\end{equation}
where $N_{(k)}^+ = \lvert I_{(k)}^+ \rvert$ and $N_{(k)}^- = \lvert I_{(k)}^- \rvert$ are the cardinalities of $ I_{(k)}^+$ and $I_{(k)}^-$, respectively, yields the binary decision-value function to be given as
\begin{equation}
    h_{(k)} \left(x^{(i)}\right) = \text{sgn} \left(\tilde{y}_{(k)}^{(i)}\right)  \frac{4N_{(k)}^+ N_{(k)}^-}{N_{(k)}^+ + N_{(k)}^-}
    \label{eqn:decisonvalue}
\end{equation}
which guarantees that the policy function in (\ref{eqn:policy_function}) is the same as the given label $y$ for all $x\in\mathcal{X}$, i.e., $y^{(i)}=\pi \left(x^{(i)}; \theta\right)$ for all $i$ where $\theta = \left\{\tilde{y}_{(k)}\right\}_{k=1}^{M}$ is the policy parameter.

\begin{proof} 
See Appendix.
\end{proof}

With the updated policy function, the agent selects an action using the $\epsilon$-greedy policy to account for the balanced exploitation and exploration:
\begin{equation}
u_t = \begin{cases} \pi(x_t; \theta) &\mbox{w.p. } 1-\epsilon \\
\text{any}~ u\in\mathcal{U}_{\text{safe}}(x_t) & \mbox{w.p. } \frac{\epsilon}{\left\|\mathcal{U}_{\text{safe}}(x_t) \right\|} \end{cases}
\label{eqn:epsilon_g}
\end{equation}
where $\epsilon$ is a probability of choosing an action from the set of safe actions. The proposed algorithm described above is summarized in Algorithm \ref{alg:rl_algorithm}.

\begin{algorithm}[t]
 \SetAlgoLined 
Initialize $\pi(x;\theta)$, $Q(x,u)~\forall x\in\mathcal{X}, u\in\mathcal{U}$ 

\Repeat{convergence}{
Initialize $x_0$ \\
\Repeat{$x_{t}\in \mathcal{X}_f$}{
1. Observe $x_t$ and take an action $u_t$ using $\epsilon$-greedy in (\ref{eqn:epsilon_g}).

2. The environment returns the next state $x_{t+1}$ and the reward $r_{t+1}$.

3. Update the $Q$-table at $(x_t,u_t)$ based on (\ref{eqn:q_update_svm}).

4. Construct a policy table from the updated $Q$-table using (\ref{eqn:labeledData}).

5. Update the policy function parameter $\theta$ using (\ref{eqn:policy_function}) and (\ref{eqn:decisonvalue}) and $t\leftarrow t+1$.}
}
 \caption{Safety-guaranteed RL based on MC-SVM}
 \label{alg:rl_algorithm}
 \end{algorithm}

We then present the theoretical results of Algorithm \ref{alg:rl_algorithm} in terms of convergence, optimality, and safety.

\textbf{Theorem 2}
\begin{enumerate}
    \item \textit{Convergence and optimality}: The $Q$-learning algorithm (\ref{eqn:q_update_svm}) converges to the optimal value function, $Q^*(x,u)$.\\
    \item \textit{Safety}: For any state $x$, the policy function $\pi(x; \theta)$ using (\ref{eqn:policy_function}) and (\ref{eqn:decisonvalue}) satisfies the constraint $C(f(x, u)) < 0$ during learning and execution.
\end{enumerate}
\begin{proof}
For the first statement, the Actor using the MC-SVM makes the proposed algorithm equivalent to $Q$-learning because the policy function is guaranteed to be the same as the given labels from the $Q$-table, as in Theorem 1. Therefore, the theorem of $Q$-learning \cite{jaakkola1994convergence} proves that the proposed algorithm converges to the optimal policy. The second statement is true by construction. In the proposed algorithm, the Critic only considers actions in the set of safe actions $\mathcal{U}_{safe}(x)$ for each state $x$. Therefore, all the actions included in the labeled dataset (\ref{eqn:labeledData}) are safe actions. From Theorem 1, the policy function ensures that the constraints are satisfied during learning and execution. 
\end{proof}

\section{Numerical Results} \label{sec:4}
\begin{figure}[t]
    \center{\includegraphics[width=135mm]
    {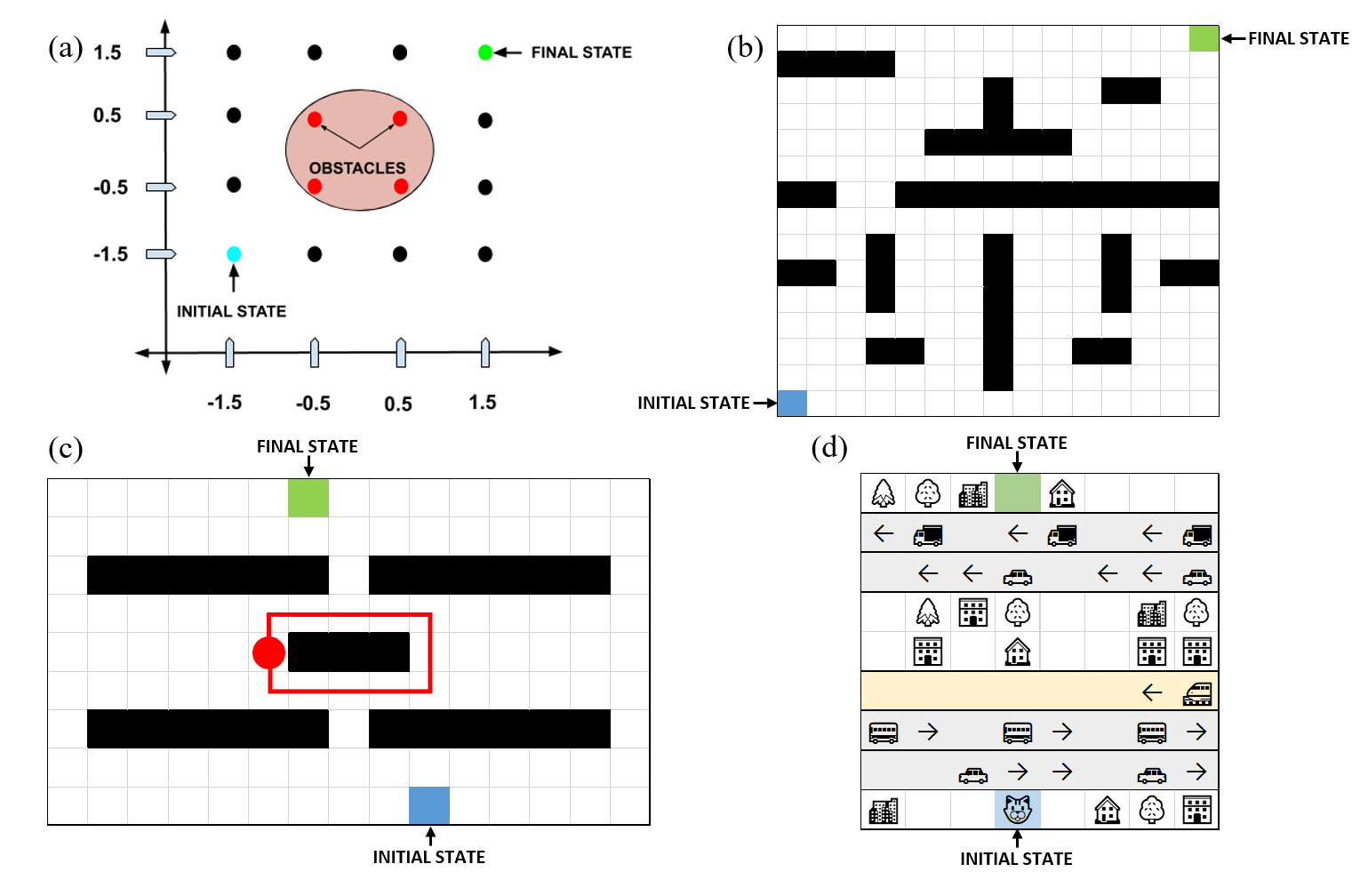}}
    \caption{Four different examples: (a) $n\times n$ grid world, (b) maze world, (c) 15x9 grid world with both fixed and moving obstacles, and (d) \emph{Crossy Road}-like game}
    \label{fig:nuexample}
\end{figure}

To demonstrate that the proposed algorithm converges to the optimal policy function while satisfying all the constraints, we applied the proposed algorithm in two domains: (1) grid world with fixed obstacles only and (2) grid world with both fixed and moving obstacles, as shown in Fig.~\ref{fig:nuexample}. The simulations were performed on a computer equipped with an Intel(R) Core(TM) i7-9750H and 16 GB of RAM running a 64-bit version of MATLAB R2016a.

\begin{figure}[h]
    \center{\includegraphics[width=130mm]
    {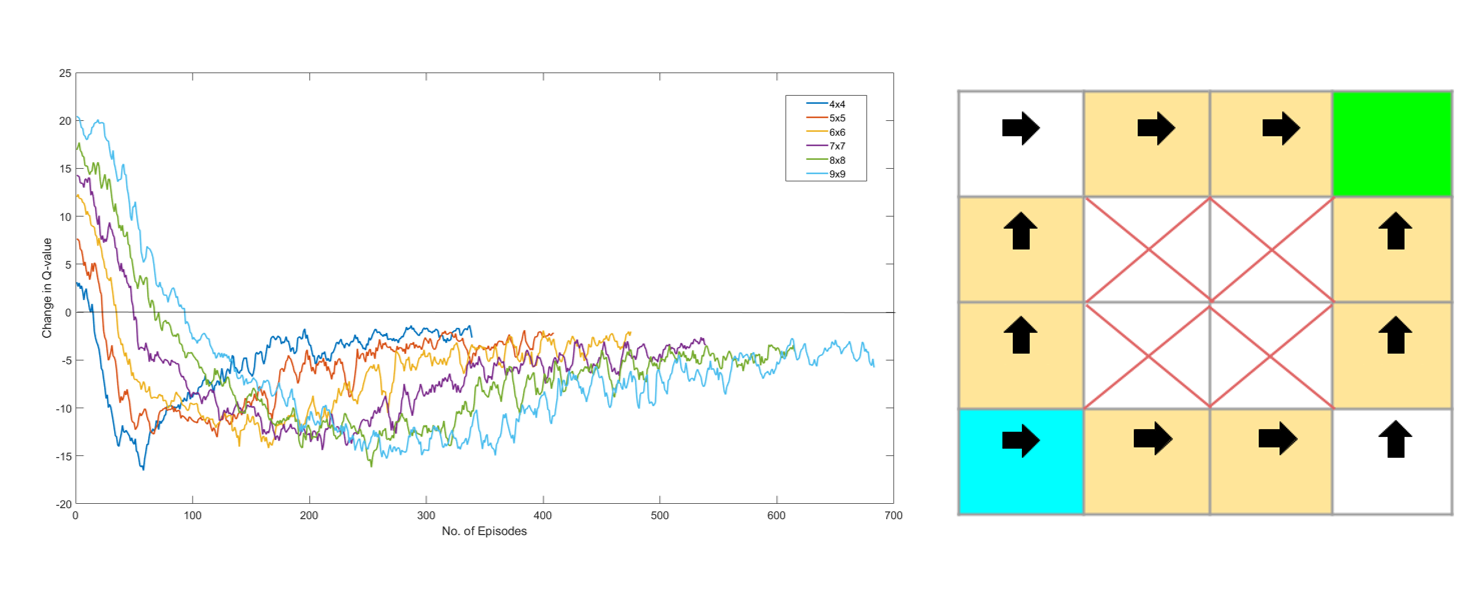}}
    \caption{The simulation results for $n \times n$ grid world: the convergence trend (for $n=4,\cdots,9$; left) and the converged policy (for $n=4$; right).}
    \label{fig:4by4}
\end{figure}

The goal of the agent is to start from the initial state (blue) and reach the final state (green) with the minimum traveled distance, while avoiding the obstacles. There are four actions available to the agent at each state: right, up, left, and down, and the state is the two dimensional position.

\begin{figure}[t]
    \center{\includegraphics[width=130mm]
    {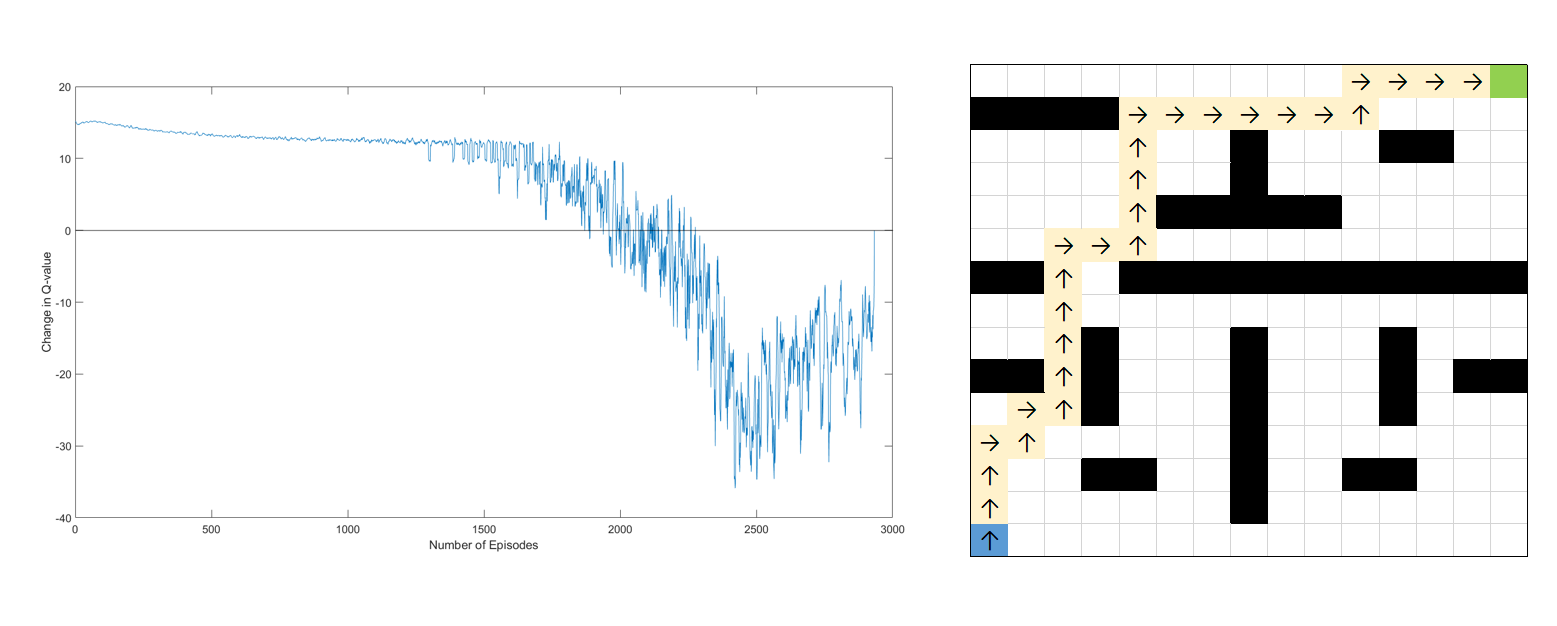}}
    \caption{The simulation results for $15 \times 15$ maze world: the convergence trend (left) and the optimal policy (right).}
    \label{fig:15by15}
\end{figure}

The fixed obstacle cases (Fig. \ref{fig:nuexample} (a) and (b)) were run with learning rate $\beta = 0.07$ and discount factor $\gamma = 1.0$. In $n \times n$ grid world, Fig. \ref{fig:4by4} (a) shows the convergence trend of the $Q$-table for the size of the grid world varying from $4 \times 4$ to $9 \times 9$, in which the vertical axis denotes the total change in the $Q$-table over an episode while the horizontal axis denotes the number of episodes required for convergence, i.e. when the change in the values of the $Q$-table is less than a threshold. In the maze world, the proposed algorithm converged within 3,000 episodes, as shown in Fig. \ref{fig:15by15} (a). In  Fig. \ref{fig:4by4} (b) and  Fig. \ref{fig:15by15} (b), it is shown that the proposed algorithm successfully converged to a safe and optimal policy.

\begin{figure}[t]
    \center{\includegraphics[width=130mm]
    {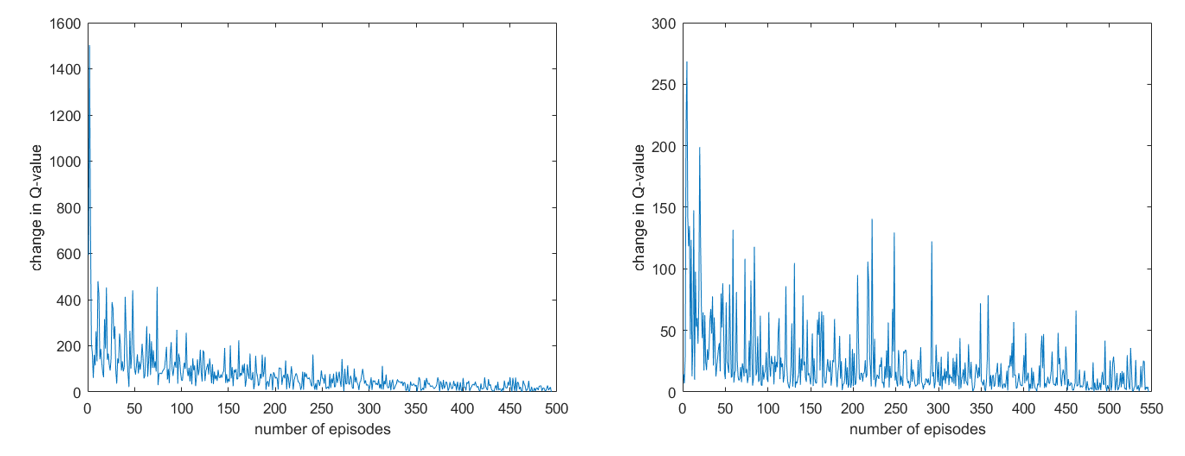}}
    \caption{The convergence trend for $15 \times 9$ grid world with moving obstacle (left) and the \emph{Crossy Road}-like game (right).}
    \label{fig:movingresult}
\end{figure}

\begin{figure}[t]
    \center{\includegraphics[width=130mm]
    {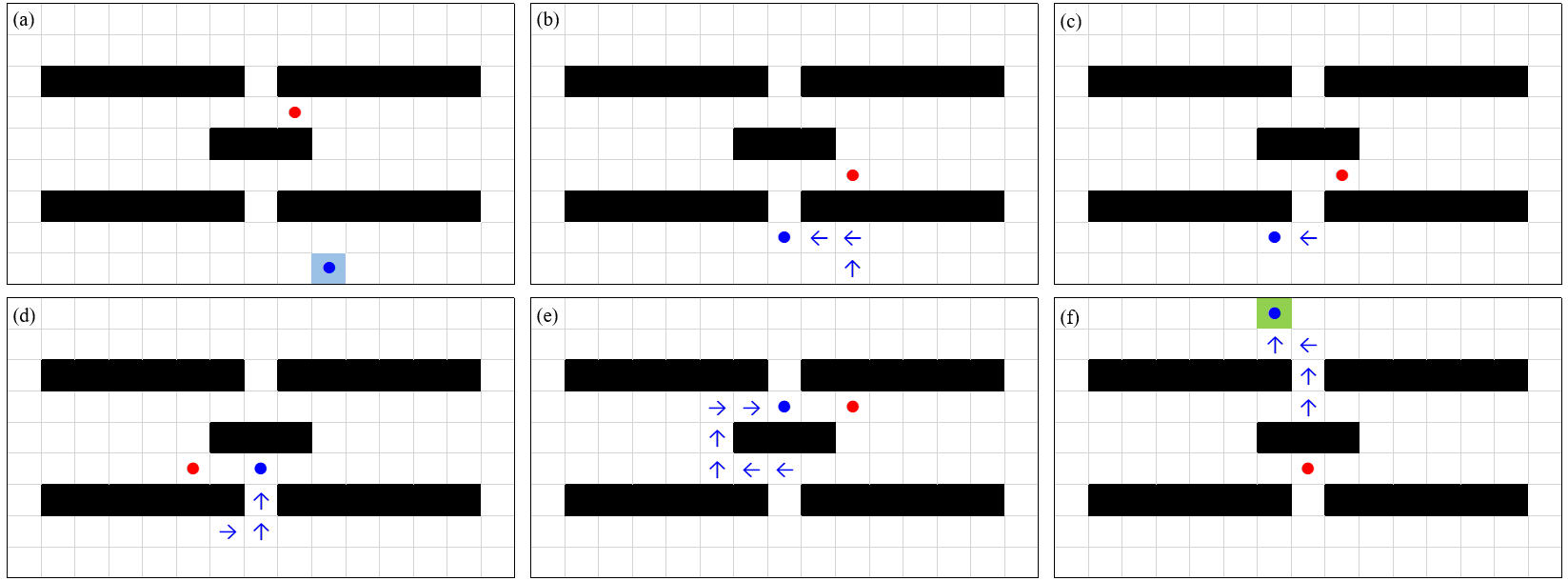}}
    \caption{The converged policy for $15 \times 9$ grid world with moving obstacle.}
    \label{fig:movingpolicy}
\end{figure}

\begin{figure}[h!h!]
    \center{\includegraphics[width=130mm]
    {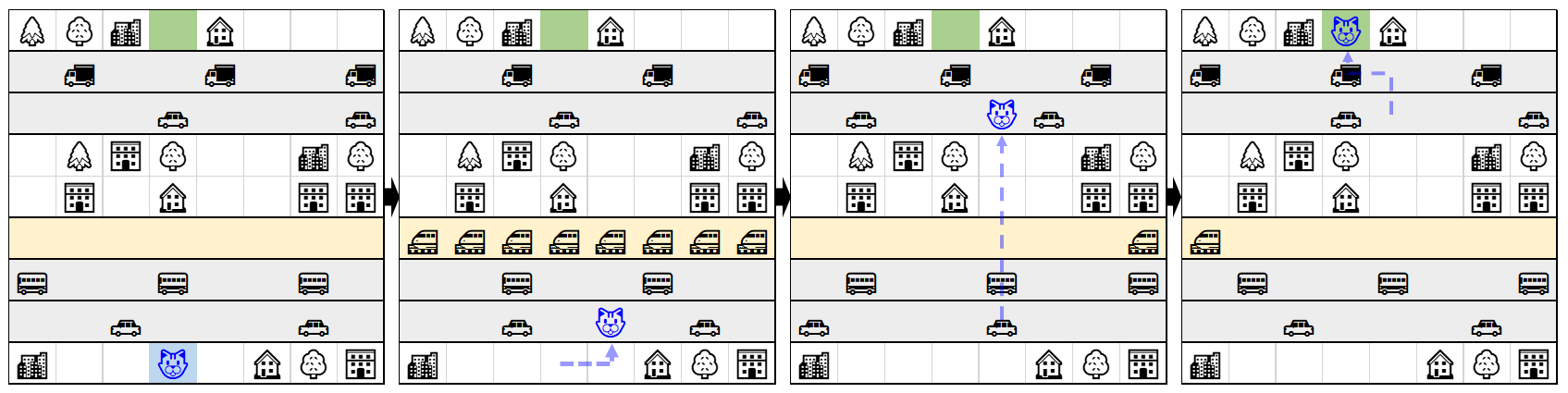}}
    \caption{The optimal policy for \emph{Crossy Road}-like game.}
    \label{fig:crossypolicy}
\end{figure}

The examples in Fig. \ref{fig:nuexample} (c) and (d) consider both fixed and moving obstacles. The state is the concatenation of the agent state and moving obstacle state. In the $15 \times 9$ grid world with a moving obstacle, there are fixed obstacles (in black) and an obstacle (in red) moving around the middle block. In the \emph{Crossy Road}-like game, there are fixed obstacles (trees, houses, and buildings) and moving obstacles (cars, trucks, and a train) with different speeds. For the both examples, Fig. \ref{fig:movingresult} shows the convergence results. As shown in Fig. \ref{fig:movingpolicy} and \ref{fig:crossypolicy}, the agent follows the optimal policy and arrives in the final state without colliding with any obstacles.

\section{Conclusions and Future Work} \label{sec:5}
In this paper, we proposed an algorithm for safety-guaranteed reinforcement learning using Multi-Class Support Vector Machine (MC-SVM), based on the Actor-Critic framework. An analytic solution was derived for the MC-SVM for the Actor's policy function, which converges to a solution guaranteed to satisfy the constraints. Additionally, the proposed formulation adheres to the $Q$-learning framework and hence also guarantees convergence to the optimal solution. The simulation results demonstrated that the final policy converges to the safe and optimal solution.

Extending the proposed algorithm to the continuous state and/or action domain is a challenging task. Training an SVM requires a finite number of states but discretizing the continuous state space is not feasible for complex problems due to the curse of dimensionality. However, there exists work which has used an SVM to learn the $Q$-value function in the continuous state space \cite{SVM_cont} by using the number of time steps in an episode as a finite measure to train the SVM instead of the states. This idea can be potentially extended to incorporate constraints in the continuous state space.

\section*{Broader Impact}
Our work is theoretical enough that there is no particular application foreseen.

\section*{Acknowledgement}
This work is partially supported by Collins Aerospace, and we would like to thanks Alex Postnikov and Josh Bertram for their support. We are also grateful to Rahul Mewada for his valuable support in implementing the proposed algorithm.
 
\small
\bibliographystyle{IEEEtran}
\bibliography{ref}

\newpage
\section*{Appendix}
For the notational simplicity, the subscript $(k)$, which represents each class of the labels, $k\in\{1,\cdots,M\}$, is omitted in this appendix. Also, we denote the kernel as $K_{ij} := K\left(x^{(i)},x^{(j)}\right)$. For the kernel as a similarity measure, it takes the maximum value, $K_{\text{max}}$, when the two arguments are the same, i.e., $i=j$. Especially for the Gaussian kernel used in this paper, the maximum value is $K_{\text{max}} = 1$ from (\ref{eqn"kernel_gauss}). We also denote $\Delta K_{ij} := K_{\text{max}}- K_{ij}$. Note that $K_{ij} = K_{ji}$, and hence $\Delta K_{ij} = \Delta K_{ji}$.

\textbf{Lemma 1} The binary decision-value function in (\ref{eqn:binary_dec}), $h\left(x^{(l)}\right)$ for $l \in \left\{1,\cdots,N\right\}$, has the same sign with $\tilde{h} \left(x^{(l)}\right)$:
\begin{equation}
    \begin{split}
        \tilde{h} \left(x^{(l)}\right) &:= h\left(x^{(l)}\right) \sum_{i=1}^N \alpha^{(i)} \\
    &= \sum_{\substack{i=1\\i\neq l}}^N \left(\alpha^{(i)}\right) ^2 \left(-2 \tilde{y}^{(i)} \right) \Delta K_{li} + \frac{1}{2} \sum_{\substack{i,j=1 \\ i,j\neq l \\ i\neq j}}^N  \alpha^{(i)} \alpha^{(j)} \left(\tilde{y}^{(i)} + \tilde{y}^{(j)}\right) \left(\Delta K_{ij} - \Delta K_{li} - \Delta K_{lj} \right)        
    \end{split}
    \label{eqn:Lemma1}
\end{equation}

\begin{proof}
We note that the offset $b$ in (\ref{eqn:binary_dec}) can be computed after the binary SVM optimization (\ref{eqn:P2}) is solved for $\alpha$ and it contains a logical expression, that is, it is computed for any $x$ with $\alpha \neq 0$. To derive an analytic solution, we first incorporate the offset term into the binary decision-value function. Let us define the \textit{pseudo-offset}, $\tilde{b}^{(i)}$, for $i=1,\cdots,N$, as
\begin{equation}
    \tilde{b}^{(i)} = \tilde{y}^{(i)} - \sum_{j=1}^N \alpha^{(j)} \tilde{y}^{(j)} K_{ij}
        \label{eqn:P_offset}
\end{equation}

Note that the offset $b$ is constant as $b = \tilde{y}^{(i)} - \sum_{j=1}^N \alpha^{(j)} \tilde{y}^{(j)} K_{ij}$ only for the \textit{support vectors}, i.e., $x^{(i)}$ with $\alpha^{(i)}\neq0$. Using this fact, we can write 
\begin{equation}
    \sum_{i=1}^N \alpha^{(i)}\tilde{b}^{(i)} = b \sum_{\substack{i=1 \\ \alpha^{(i)}\neq 0}}^N \alpha^{(i)} =b \sum_{i=1}^N \alpha^{(i)}
    \label{eqn:offset}
\end{equation}
Multiplying (\ref{eqn:binary_dec}) by $\sum_{i=1}^N \alpha^{(i)}$, we get
\begin{equation}
\begin{split}
    \tilde{h} \left(x^{(l)}\right) = h\left(x^{(l)}\right) \sum_{i=1}^N \alpha^{(i)} 
    & \stackrel{}{=} \sum_{i,j=1}^N {\alpha^{(i)}\alpha^{(j)}}\Tilde{y}^{(i)} K_{li} +b \sum_{i=1}^N \alpha^{(i)} \\
    & \stackrel{(\ref{eqn:offset})}{=} \sum_{i,j=1}^N {\alpha^{(i)}\alpha^{(j)}}\Tilde{y}^{(i)} K_{li} +\sum_{i=1}^N \alpha^{(i)}\tilde{b}^{(i)} \\
    & \stackrel{(\ref{eqn:P_offset})}{=} \sum_{i,j=1}^N {\alpha^{(i)}\alpha^{(j)}}\Tilde{y}^{(i)} K_{li} +\sum_{i=1}^N \alpha^{(i)}\tilde{y}^{(i)}
    -\sum_{i,j=1}^N {\alpha^{(i)}\alpha^{(j)}}\Tilde{y}^{(i)} K_{ij}
        \label{eqn:htilde}
\end{split}
\end{equation}
    
Since $\sum_{i=1}^N \alpha^{(i)} \ge 0$ because $\alpha^{(i)} \in [0, U_b]$ as in (\ref{eqn:P2}), the signs of $\tilde{h}$ and $h$ are the same. The second term in (\ref{eqn:htilde}) is removed by incorporating the equality constraint given in (\ref{eqn:P2}), and hence,
\begin{equation}
     \tilde{h} \left(x^{(l)}\right) = \sum_{i,j=1}^N {\alpha^{(i)}\alpha^{(j)}}\Tilde{y}^{(i)} \left(K_{li} - K_{ij} \right) 
     \label{eqn:quadratic}
\end{equation}

By (\ref{eqn:quadratic}), we now have the offset term incorporated into the binary decision-value function. By further expanding the terms,
\begin{equation}
    \tilde{h}\left(x^{(l)}\right) = \sum_{\substack{i=1 \\ i\neq l}}^N \alpha^{(i)} \left(\alpha^{(l)}\tilde{y}^{(l)} - \alpha^{(i)}\tilde{y}^{(i)}\right) \left(K_{\text{max}} - K_{li}\right) + \sum_{\substack{i,j=1\\i,j\neq l \\ i\neq j}}^N \alpha^{(i)}\alpha^{(j)} \tilde{y}^{(i)} \left(K_{li} - K_{ij}\right)
\end{equation}
By subtracting and adding the terms,
\begin{equation}
\begin{split}
    \tilde{h}\left(x^{(l)}\right) = &\sum_{\substack{i=1 \\ i\neq l}}^N \alpha^{(i)} \left(\alpha^{(l)}\tilde{y}^{(l)} - \alpha^{(i)}\tilde{y}^{(i)} - \tilde{y}^{(i)} \sum_{\substack{j=1\\j\neq l\\j\neq i}}^N \alpha^{(j)} + \tilde{y}^{(i)} \sum_{\substack{j=1\\j\neq l\\j\neq i}}^N \alpha^{(j)} \right) \left(K_{\text{max}} - K_{li}\right) \\ 
    &+ \sum_{\substack{i,j=1\\i,j\neq l \\ i\neq j}}^N \alpha^{(i)}\alpha^{(j)} \tilde{y}^{(i)} \left(K_{li} - K_{ij}\right) \\
    = & \sum_{\substack{i=1\\i\neq l}}^N \alpha^{(i)} \left( \alpha^{(l)} \tilde{y}^{(l)} - \tilde{y}^{(i)} \sum_{\substack{j=1 \\ j\neq l}}^N \alpha^{(j)} \right) \Delta K_{li} + \sum_{\substack{i,j=1 \\ i,j\neq l \\ i\neq j}}^N \alpha^{(i)} \alpha^{(j)} \tilde{y}^{(i)} \Delta K_{ij}
\end{split}
\end{equation}

Using the equality constraint, we have $\alpha^{(l)}\tilde{y}^{(l)}  = - \sum_{\substack{i=1 \\ i \neq l}}^N \tilde{y}^{(i)} \alpha^{(i)}$, which yields

\begin{equation}
    \begin{split}
    \tilde{h}\left(x^{(l)}\right) 
    =&\sum_{\substack{i=1 \\ i \neq l}}^N \alpha^{(i)} \left( - \sum_{\substack{j=1\\j\neq l}}^N \tilde{y}^{(j)}\alpha^{(j)} - \tilde{y}^{(i)} \sum_{\substack{j=1 \\ j\neq l}}^N \alpha^{(j)}\right) \Delta K_{li} + \sum_{\substack{i,j=1\\ i,j\neq l \\ i\neq j}}^N \alpha^{(i)}\alpha^{(j)}\tilde{y}^{(i)} \Delta K_{ij}\\
    = &\sum_{\substack{i,j=1 \\ i,j\neq l}}^N \alpha^{(i)}\alpha^{(j)} \left(-\tilde{y}^{(i)}-\tilde{y}^{(j)}\right) \Delta K_{li} + \sum_{\substack{i,j=1\\ i,j\neq l \\ i\neq j}}^N \alpha^{(i)}\alpha^{(j)}\tilde{y}^{(i)} \Delta K_{ij}\\
    =& \sum_{\substack{i=1\\i\neq l}}^N \left(\alpha^{(i)}\right)^2 \left(-2\tilde{y}^{(i)}\right) \Delta K_{li} + \sum_{\substack{i,j=1 \\ i,j\neq l \\ i\neq j}}^N \alpha^{(i)} \alpha^{(j)} \left(-\tilde{y}^{(i)} \right) \Delta K_{li} \\
    &+ \sum_{\substack{i,j=1 \\ i,j\neq l \\ i\neq j}}^N \alpha^{(i)} \alpha^{(j)} \left(-\tilde{y}^{(i)} \right) \Delta K_{lj} + \sum_{\substack{i,j=1\\ i,j\neq l \\ i\neq j}}^N \alpha^{(i)}\alpha^{(j)}\tilde{y}^{(i)} \Delta K_{ij}\\
    = & \sum_{\substack{i=1\\i\neq l}}^N \left(\alpha^{(i)}\right)^2 \left(-2\tilde{y}^{(i)}\right) \Delta K_{li} + \sum_{\substack{i,j=1 \\ i,j\neq l\\ i\neq j}}^N \alpha^{(i)} \alpha^{(j)} \tilde{y}^{(i)} \left( \Delta K_{ij} - \Delta K_{li} - \Delta K_{lj} \right) \\
    = & \sum_{\substack{i=1\\i\neq l}}^N \left(\alpha^{(i)}\right)^2 \left(-2\tilde{y}^{(i)}\right) \Delta K_{li} + \frac{1}{2} \sum_{\substack{i,j=1 \\i,j\neq l\\i\neq j}} ^N \alpha^{(i)} \alpha^{(j)} \left( \tilde{y}^{(i)} + \tilde{y}^{(j)} \right) \left( \Delta K_{ij} - \Delta K_{li} - \Delta K_{lj} \right)
    \end{split}
\end{equation}
\end{proof}

Since $\tilde{h}$ and $h$ operate equivalently in the process of the MC-SVM, we denote $\tilde{h}$ as $h$ afterwards.

\textbf{Theorem 1} Let the Lagrangian multipliers $\alpha^{(i)}$ corresponding to $I^+$ and $I^-$ be $\alpha^+$ and $\alpha^-$, respectively. For $\eta\rightarrow\infty$, the analytic solution
\begin{equation}
\alpha^+ = \frac{2N^-}{N} \text{ and } \alpha^- = \frac{2N^+}{N} \tag{\ref{eqn:analytic}}
\end{equation}
where $N^+ = \lvert I^+ \rvert$ and $N^- = \lvert I^- \rvert$ are the cardinalities of $ I^+$ and $I^-$, respectively, yields the binary decision-value function to be given as
\begin{equation}
    h \left(x^{(i)}\right) = \text{sgn} \left(\tilde{y}^{(i)}\right)  \frac{4N^+ N^-}{N^+ + N^-} \tag{\ref{eqn:decisonvalue}}
\end{equation}
which guarantees that the policy function in (\ref{eqn:policy_function}) is the same as the given label $y$ for all $x\in\mathcal{X}$, i.e., $y^{(i)}=\pi \left(x^{(i)}; \theta\right)$ for all $i$ where $\theta = \left\{\tilde{y_{(k)}}\right\}_{k=1}^{M}$ is the policy parameter.

\begin{proof}
Assume that $N^+ \neq 0 $ and $N^- \neq 0$, that is, there is at least one positive or negative label in the binary dataset. 
For the Gaussian kernel, if $\eta \rightarrow \infty$, $K_{ij}$ becomes zero if $i\neq j$, and hence $\Delta K_{ij} = K_{\text{max}}=1$.If $i=j$, $\Delta K_{ij} = K_{\text{max}}-K_{ii} = K_{\text{max}}-K_{\text{max}} = 0$. With this fact and $\alpha^+$ and $\alpha^-$ given in Theorem 1, the objective function in (\ref{eqn:P2}) becomes

\begin{equation}
    W = \frac{1}{2} \left\{ N^+ \left(\alpha^+\right)^2 + N^- \left(\alpha^-\right)^2 \right\} - \left( N^+ \alpha^+ + N^- \alpha^- \right)
    \label{eq:objectiveF}
\end{equation}

From the equality constraint in (\ref{eqn:P2}), we have $N^+ \alpha^+ = N^- \alpha^-$. Plugging $\alpha^+ = \frac{N^-}{N^+} \alpha^-$ into (\ref{eq:objectiveF}) and taking the derivative of $W$ with respect to $\alpha^-$ yields the analytic solutions in (\ref{eqn:analytic}).

For $\eta \rightarrow \infty$, the binary decision-value function in (\ref{eqn:Lemma1}) becomes

\begin{equation}
     \tilde{h}\left(x^{(l)}\right) = 
     -\sum_{\substack{i=1\\i\neq l}}^N \left(\alpha^{(i)}\right)^2 \left(2\tilde{y}^{(i)}\right)  
     - \frac{1}{2} \sum_{\substack{i,j=1 \\i,j\neq l\\i\neq j}} ^N \alpha^{(i)} \alpha^{(j)} \left( \tilde{y}^{(i)} + \tilde{y}^{(j)} \right)
\end{equation}

Suppose that the label corresponding to the state $x^{(l)}$ is given as positive, i.e., $\tilde{y}^{(l)}=+1$. Then, for the other $N-1$ states, there are $N^+ -1$ positive labels and $N^-$ negative labels, which yields
\begin{equation}
\begin{split}
    \tilde{h}\left(x^{(l)}\right) = 
    &-\left(\alpha^+\right)^2 (+2) \left(N^+ - 1 \right) 
    -\left(\alpha^-\right)^2 (-2) \left(N^- \right) \\
    &- \frac{1}{2} \left(\alpha^+\right)^2 (+2) P_{N^+ -1, 2}
    - \frac{1}{2} \left(\alpha^-\right)^2 (-2) P_{N^-, 2}\\
    =& - N^+ (N^+ - 1) \left(\alpha^+ \right)^2 + N^- (N^- + 1) \left(\alpha^- \right)^2
\end{split}
\label{eq:case1}
\end{equation}
where $P_{n,r}$ represents the $r$-permutations of $n$, computed as $P_{n,r} = \frac{n!}{(n-r)!}$. Plugging (\ref{eqn:analytic}) into (\ref{eq:case1}) leads to
\begin{equation}
    \tilde{h}\left(x^{(l)}\right) = \frac{4N^+ N^-}{N^+ + N^-}    
    \label{eqn:positive}
\end{equation}
Similarly, if $\tilde{y}^{(l)}=-1$, we have
\begin{equation}
    \tilde{h}\left(x^{(l)}\right) = -\frac{4N^+ N^-}{N^+ + N^-}    
    \label{eqn:negative}
\end{equation}
By combining (\ref{eqn:positive}) and (\ref{eqn:negative}), we obtain the binary decision-value function as in (\ref{eqn:decisonvalue}). 


For the state $x^{(l)}$, suppose that its multi-class label is given as $y^{(l)}=m \in \{1,\cdots, M\}$. For each $k\in \{ 1, \cdots, M\}$, by (\ref{eqn:multilabels}), the binary labels corresponding to $x^{(l)}$ are constructed as $\tilde{y}^{(l)}_{(m)} = +1$ and $\tilde{y}^{(l)}_{(k)} = -1$ for all $k\neq m$. Therefore, from (\ref{eqn:decisonvalue}), the binary decision-value function is positive only for $k=m$ and negative for all $k\neq m$. From (\ref{eqn:policy_function}), the policy function is then given as $\pi\left(x^{(l)};\theta \right) = m$, which is the same as the given label $y^{(l)} = m$.

As a final note, consider the case where either $N_{(\tilde{k})}^+$ or $N_{(\tilde{k})}^-$ for a label (or an action) $\tilde{k}\in\{1,\cdots,M\}$ is zero during a certain step of Algorithm \ref{alg:rl_algorithm}. 
Consider a case when the action $\tilde{k}$ does not appear for any of the states, which leads to $N_{(\tilde{k})}^+ = 0$. This makes the corresponding binary decision-value function zero, i.e., $h_{(\tilde{k})}\left(x^{(i)}\right) = 0$, for all $i\in\{1,\cdots,N\}$. However, since each state $x$ must be assigned to an action other than $\tilde{k}$, say $m \neq \tilde{k}$, it has the only one positive decision-value $h_{(m)} (x) > 0$ and the negative ones $h_{(k)} (x) <0$ for all $k\neq \tilde{k}, m$, thereby (\ref{eqn:policy_function}) being equal to $m$. This holds whenever at least two of $M$ actions are selected by the states. 
\end{proof}

\textbf{Remark 1}. If all the states are assigned to the same action $\tilde{k}$, then $N_{(\tilde{k})}^+ = 0$ and $N_{(k)}^- = 0$ for all $k\neq \tilde{k}$, which leads to $h_{(k)} \left(x^{(i)}\right) =0 $ for all $i$ and $k$. This may be due to the current policy function or the random exploration , as in (\ref{eqn:epsilon_g}). In this case, the agent should go to Line 5 of Algorithm \ref{alg:rl_algorithm} until at least two actions are selected by the states in the policy table, in Line 8 of Algorithm \ref{alg:rl_algorithm}.
\end{document}